\documentclass[sigconf]{acmart}
\usepackage{booktabs,tabularx}
\usepackage{enumitem}
\usepackage{multirow}
\usepackage{microtype}
\usepackage{subcaption} 

\usepackage{amsmath,amsfonts}
\theoremstyle{definition}
\newtheorem{definition}{Definition}[section]
\newtheorem*{problem}{Problem} 
\newtheorem*{example}{Example} 

\usepackage{pifont} 
\newcommand{\cmark}{\ding{51}} 
\newcommand{\xmark}{\ding{55}} 

\AtBeginDocument{%
  }

\copyrightyear{2026}
\acmYear{2026}
\setcopyright{cc}
\setcctype{by}
\acmConference[WWW '26] {Proceedings of the ACM Web Conference 2026}{April 13--17, 2026}{Dubai, United Arab Emirates.}
\acmBooktitle{Proceedings of the ACM Web Conference 2026 (WWW '26), April 13--17, 2026, Dubai, United Arab Emirates}
\acmISBN{979-8-4007-2307-0/2026/04}
\acmDOI{10.1145/3774904.3792710}
\usepackage{multirow}
\begin{document}

\title{HyperRAG: Reasoning N-ary Facts over Hypergraphs for Retrieval Augmented Generation}


\author{Wen-Sheng Lien}
\orcid{0009-0004-1107-5590}
\affiliation{%
  \institution{National Yang Ming Chiao Tung University}
  \city{Hsinchu}
  \country{Taiwan}
}
\email{vincentlien.ii13@nycu.edu.tw}

\author{Yu-Kai Chan}
\orcid{0009-0005-0468-2525}
\affiliation{%
  \institution{National Yang Ming Chiao Tung University}
  \city{Hsinchu}
  \country{Taiwan}
}
\email{ctw33888.ee13@nycu.edu.tw}

\author{Hao-Lung Hsiao}
\orcid{0009-0002-2120-4881}
\affiliation{%
  \institution{National Yang Ming Chiao Tung University}
  \city{Hsinchu}
  \country{Taiwan}
}
\email{hlhsiao.cs13@nycu.edu.tw}

\author{Bo-Kai Ruan}
\orcid{0000-0002-9847-3628}
\affiliation{%
  \institution{National Yang Ming Chiao Tung University}
  \city{Hsinchu}
  \country{Taiwan}
}
\email{bkruan.ee11@nycu.edu.tw}

\author{Meng-Fen Chiang}
\orcid{0009-0008-8385-0380}
\affiliation{%
  \institution{National Yang Ming Chiao Tung University}
  \city{Hsinchu}
  \country{Taiwan}
}
\email{meng.chiang@nycu.edu.tw}

\author{Chien-An Chen}
\orcid{0009-0007-5142-5243}
\affiliation{%
  \institution{E.SUN Bank}
  \city{Taipei}
  \country{Taiwan}
}
\email{lukechen-15953@esunbank.com}

\author{Yi-Ren Yeh}
\orcid{0000-0003-4264-523X}
\affiliation{%
  \institution{National Kaohsiung Normal University}
  \city{Kaohsiung}
  \country{Taiwan}
}
\email{yryeh@nknu.edu.tw}

\author{Hong-Han Shuai}
\orcid{0000-0003-2216-077X}
\affiliation{%
  \institution{National Yang Ming Chiao Tung University}
  \city{Hsinchu}
  \country{Taiwan}
}
\email{hhshuai@nycu.edu.tw}

\renewcommand{\shortauthors}{Wen-Sheng Lien et al.}

\begin{abstract}
Graph-based Retrieval-Augmented Generation (RAG) typically operates on binary Knowledge Graphs (KGs). However, decomposing complex facts into binary triples often leads to semantic fragmentation and longer reasoning paths, increasing the risk of retrieval drift and computational overhead. In contrast, \(n\)-ary hypergraphs preserve high-order relational integrity, enabling shallower and more semantically cohesive inference. To exploit this topology, we propose \textbf{HyperRAG}, a framework tailored for \(n\)-ary hypergraphs featuring two complementary retrieval paradigms:
(i) HyperRetriever learns structural-semantic reasoning over \(n\)-ary facts to construct query-conditioned relational chains. It enables accurate factual tracking, adaptive high-order traversal, and interpretable multi-hop reasoning under context constraints.
(ii) HyperMemory leverages the LLM’s parametric memory to guide beam search, dynamically scoring \(n\)-ary facts and entities for query-aware path expansion.
Extensive evaluations on WikiTopics (11 closed-domain datasets) and three open-domain QA benchmarks (HotpotQA, MuSiQue, and 2WikiMultiHopQA) validate HyperRAG’s effectiveness. HyperRetriever achieves the highest answer accuracy overall, with average gains of 2.95\% in MRR and 1.23\% in Hits@10 over the strongest baseline.
Qualitative analysis further shows that HyperRetriever bridges reasoning gaps through adaptive and interpretable \(n\)-ary chain construction, benefiting both open and closed-domain QA. Our codes are publicly available at \href{https://github.com/Vincent-Lien/HyperRAG.git}{https://github.com/Vincent-Lien/HyperRAG.git}.
\end{abstract}




%
\maketitle

\section{Introduction}\label{sec:intro}

\begin{figure}[t]
    \centering
    \captionsetup{type=figure}
    \includegraphics[width=0.9\linewidth]{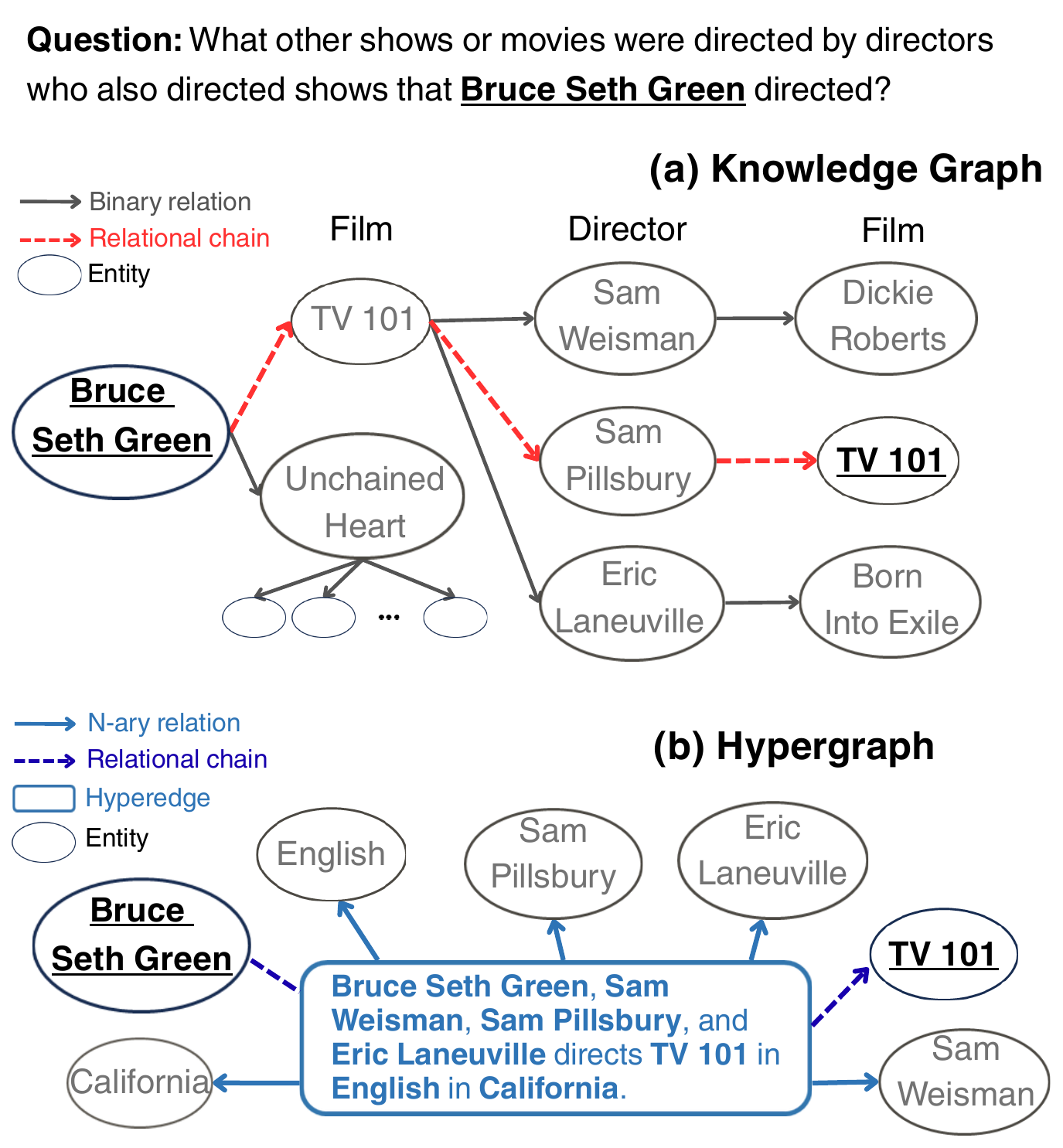}
    \captionof{figure}{Structural Comparison of (a) Knowledge Graphs and (b) Hypergraphs. For a given question $q$, (a) requires 3-hop reasoning over binary facts, while (b) enables single-hop inference via an \(n\)-ary relational fact, yielding a more compact and expressive multi-entity representation.}\label{fig:motivation}
\end{figure}

Retrieval-Augmented Generation (RAG) has established itself as a critical mechanism for augmenting Large Language Models (LLMs) with non-parametric external knowledge during inference~\cite{gao2024,Huang_2025,lee2025nvembed,karpukhin-etal-2020-dense}. By dynamically retrieving verifiable information from external corpora without the need for extensive fine-tuning, RAG effectively mitigates intrinsic LLM limitations such as hallucinations and temporal obsolescence. This paradigm has proven particularly transformative for knowledge-intensive tasks, including open-domain question answering (QA), fact verification, and complex information extraction, driving significant innovation across both academia and industry.

Current RAG methodologies broadly fall into three categories: document-based, graph-based, and hybrid approaches. Document-based methods utilize dense vector retrieval to match queries with textual segments, offering scalability but often failing to capture complex structural dependencies~\cite{chen-etal-2024-m3,moreira2024}. Conversely, graph-based methods leverage Knowledge Graphs (KGs) to explicitly model relationships, enabling multi-hop reasoning over structured data~\cite{peng2024graphrag,han2025graphrag}. Hybrid approaches attempt to bridge these paradigms, balancing comprehensiveness with efficiency. However, despite the reasoning potential of graph-based methods, the prevailing reliance on binary KGs presents fundamental topological limitations.


Traditional graph-based RAG methods predominantly rely on binary knowledge graphs, which suffer from notable limitations when applied to closed-domain question-answering scenarios. 
Specifically, binary KG approaches encounter two fundamental structural limitations. 
First, \textbf{Semantic Fragmentation} arises because binary relations limit the expressiveness required to capture complex multi-entity interactions, forcing the decomposition of holistic facts into disjoint triples that fail to represent intricate semantic nuances.  
Second, this fragmentation leads to \textbf{Path Explosion}, where conventional approaches incur significant computational costs due to the need for deep traversals over the vast binary relation space to reconnect these facts, enabling error propagation and undermining real-world practicality~\cite{sun2024thinkongraph,jiang-etal-2023-active}. 
To address these limitations, recent work advocates hypergraphs for structured retrieval in RAG. Hypergraphs natively encode higher-order (\(n\)-ary) relations that bind multiple entities and roles, providing a richer semantic substrate than binary graphs~\cite{text2nkg2024}. 
As illustrated in Figure~\ref{fig:motivation}, the Path Explosion issue is evident when answering a question grounded on the topic entity ``Bruce Seth Green,'' which requires a 3-hop binary traversal on a standard KG. 
In contrast, this reduces to a single hop through an $n$-ary relation in a hypergraph, yielding a more compact representation. 
Hypergraphs enable the direct modeling of higher-order relational chains, effectively mitigating Semantic Fragmentation and reducing the reasoning steps required to capture complex dependencies.

Motivated by these insights, we introduce \textbf{HyperRAG}, an innovative retrieval-augmented generation framework designed explicitly for reasoning over \(n\)-ary hypergraphs. HyperRAG integrates two novel adaptive retrieval variants: (i) \textit{HyperRetriever}, which uses a multilayer perceptron (MLP) to fuse structural and semantic embeddings, constructing query-conditioned relational chains that enable accurate and interpretable evidence aggregation within context and token constraints; and (ii) \textit{HyperMemory}, which leverages the parametric memory of an LLM to guide beam search, dynamically scoring \(n\)-ary facts and entities for query-adaptive path expansion. By combining higher-order reasoning with shallower yet more expressive chains that locate key evidence without multi-hop traversal. Replacement of the \(n\)-ary structure with a binary reduces the average MRR from \(36.45\%\) to \(34.15\%\) and the average Hits@10 from \(40.59\%\) to \(36.82\%\) (Table~3), indicating gains in response quality.

Our key contributions are summarized as follows.
\begin{itemize}[leftmargin=*]
    \item We propose HyperRAG, a pioneering framework that shifts the graph-RAG paradigm from binary triples to $n$-ary hypergraphs, tackling the issues of semantic fragmentation and path explosion.
    \item We introduce HyperRetriever, a trainable MLP-based retrieval module that fuses structural and semantic signals to extract precise, interpretable evidence chains with low latency.
    \item We develop HyperMemory, a synergistic retrieval approach that utilizes LLM parametric knowledge to guide symbolic beam search over hypergraphs for complex query adaptive reasoning.
    \item Extensive evaluation across closed-domain and open-domain benchmarks demonstrates that HyperRAG consistently outperforms strong baselines, offering a superior trade-off between retrieval accuracy, reasoning interpretability, and system latency.
\end{itemize}

\section{Preliminaries}\label{sec:pre}

\subsection{Background}
\begin{definition}[\(n\)-ary Relational Knowledge Graph]
An \(n\)-ary relational knowledge graph, or hypergraph, represents relational facts involving two or more entities and one or more relations. Formally, following the definition in \cite{NIPS2006_dff8e9c2}, a hypergraph is defined as $\mathcal{G} = (\mathcal{E}$, $\mathcal{R}$, $\mathcal{F}$), where $\mathcal{E}$ denotes the set of entities, $\mathcal{R}$ denotes the set of relations, and $\mathcal{F}$ the set of \(n\)-ary relational facts (hyperedges). Each \(n\)-ary fact $f^n \in \mathcal{F}$, which consists of two or more entities, is represented as: $f^n=\{e_i\}_{i=1}^n$, where $\{e_i\}_{i=1}^n \subseteq \mathcal{E}$ is a set of $n$ entities with $n\ge2$. 
\end{definition}

Unlike binary knowledge graphs, \(n\)-ary representation inherently captures higher-order relational dependencies among multiple entities. \(n\)-ary relations cannot be faithfully decomposed into combinations of binary relations without losing structural integrity or introducing ambiguity in semantic interpretation \cite{Silberschatz2010,Abiteboul1995,Fagin1977}. We formalize faithful reduction and show that any straightforward binary scheme violates at least one of: (i) recoverability of the original tuples, (ii) role preservation, or (iii) multiplicity of co-participations.
Please refer to Appendix~\ref{ap:reduction} for more details on the recoveryability of role-preserving hypergraph reduction, roles, and multiplicity.

\subsection{Problem Formulation}\label{sec:problem}

\begin{problem}[Hypergraph-based RAG]
Given a question $q$, a hypergraph $\mathcal{G}$ representing \(n\)-ary relational structures, and a collection of source documents $\mathcal{D}$, the goal of hypergraph-based retrieval-augmented generation (RAG) is to generate faithful and contextually grounded answers $a$ by leveraging salient multi-hop relational chains from $\mathcal{G}$ and extracting relevant textual evidence from $\mathcal{D}$.
\end{problem}

\noindent \textbf{Complexity: Native $n$-ary Hypergraph Retrieval. }
Let $N_e{=}|\mathcal{E}|$, $N_f{=}|\mathcal{F}|$, and $\bar{n}$ be the average arity.
A query binds $k$ role-typed arguments,
$q=\{(r_i{:}a_i)\}_{i=1}^k$, and asks for the remaining $n{-}k$ roles.
We maintain sorted posting lists over role incidences,
$\mathcal{P}(r{:}a)=\{f\in\mathcal{F}:(r{:}a)\in f\}$, with length $d(r{:}a)$.
To answer $q$, the \(n\)-ary based retriever intersects the $k$ posting lists \emph{by hyperedge IDs} and reads the missing roles from each surviving hyperedge. Let $n^\star$ be the (max/avg) arity among matches. The running time is given by:
\begin{equation}
T_{\text{HYP}}(q)
= \mathcal{O}\!\Big(\sum_{i=1}^k d(r_i{:}a_i)\;+\;\texttt{out}\Big),
\end{equation}
where \texttt{out} is the number of matching facts.
In typical schemas, the relation arity is often bounded by a small constant (e.g., triadic, $n\!\le\!3$). As a result, for each match the retriever touches exactly one hyperedge record to materialize the unbound roles, yielding \emph{per-output} overhead $\mathcal{O}(1)$.

\noindent \textbf{Complexity: Standard Binary KG Retrieval. }
Suppose each $n$-ary fact $f$ is reified as an event node $e_f$ with $n$ role-typed
binary edges (e.g., $\textsf{role}_j(e_f,a_j)$).
For each binding $(r_i{:}a_i)$, use the list of event IDs posted
$\mathcal{P}_{\text{event}}(r_i{:}a_i)$ and intersect the $k$ lists to obtain candidate
events to mirror the hypergraph intersection. For each surviving $e_f$, follow its remaining $(n-k)$ role-edges to materialize unbound arguments. Let $d_{\text{event}}(r{:}a)=|\mathcal{P}_{\text{event}}(r{:}a)|$ and let $n^\star$ be the (max/avg) arity over matches. The running time is given by:
\begin{align}
T_{\text{BIN}}(q)
&= \mathcal{O}\!\Big(\sum_{i=1}^k d_{\text{event}}(r_i{:}a_i)
\;+\;
\texttt{out}\cdot (n^\star-k)\Big).
\end{align}
Under a schema-bounded arity, the \emph{per-result} overhead is up to $\bar{n}$ role lookups to materialize the remaining arguments. In contrast, the hypergraph returns them from a single record.

\noindent \textbf{Complexity Gap. }
In a native hypergraph, all arguments of an \(n\)-ary fact co-reside in a \emph{single} hyperedge record, thus materializing a hit, is one read, i.e., $\mathcal{O}(1)$ per result under bounded arity.
In contrast, in an event-reified binary KG, the fact is split across $n$ role-typed edges, reachable only via the intermediate event node $e_f$. As a result, materializing requires up to $(n-k)$ pointer chases, yielding $\texttt{out}\cdot\bar{n}$ term, and usually incurs extra indirections/cache misses.

\begin{figure}[t]
    \centering
    \includegraphics[width=\linewidth]{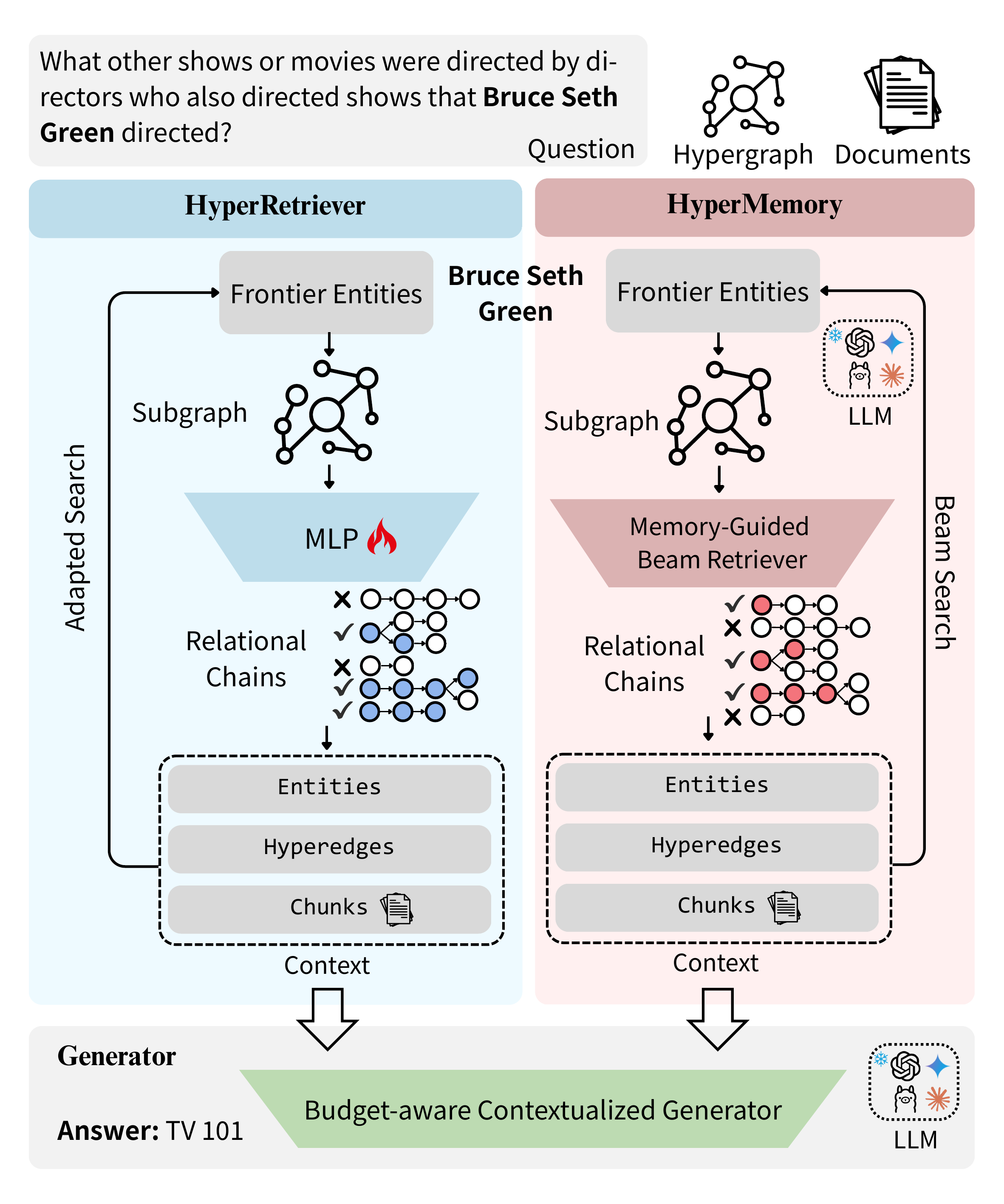}
    \caption{The overall framework of HyperRAG.}
    \label{fig:framework}
\end{figure}

\section{Methodology}\label{sec:method}
We propose \textbf{HyperRAG}, a novel framework that enhances answer fidelity by integrating reasoning over condensed $n$-ary relational facts with textual evidence. As depicted in Figure~\ref{fig:framework}, HyperRAG features two retrieval paradigms:
(i) \textbf{HyperRetriever}, which performs adaptive structural-semantic traversal to build interpretable, query-conditioned relational chains;
(ii) \textbf{HyperMemory}, which utilizes the parametric knowledge of the LLM to guide symbolic beam search. 
Both variants ground the generation process in hypergraph structures, ensuring faithful and accurate multi-hop reasoning.

\subsection{HyperRetriever: Relational Chains Learning}
The motivation behind learning to extract fine-grained \(n\)-ary relational chains over hypergraph structures stems from two key challenges: (i) the well-documented tendency of LLMs to hallucinate factual content and (ii) the vast combinatorial search space of hypergraphs under limited token and context budgets \cite{hypergraphrag2025}.
To mitigate these challenges, we introduce a lightweight yet expressive retriever that integrates structural and semantic cues to rank salient \(n\)-ary facts aligned with query intent. 

\subsubsection{Topic Entity Extraction.}
The purpose of obtaining the topic entity is to ground the query semantics onto hypergraphs $\mathcal{G}$.
Formally, given a query \(q\), we request an LLM with prompt \(p_{\text{topic}}\) to identify a set of topic entities that appear in $q$ in an LLM as follows:
\[
\mathcal{E}_q \;=\; \mathrm{LLM}\bigl(p_{\text{topic}}, q\bigr),
\]
where \(\mathcal{E}_q\) denotes the set of extracted entities in the query $q$. 

\subsubsection{Hyperedge Retrieval and Triple Formation.}
For each extracted topic entity \(e_s \in \mathcal{E}_q \), we retrieve its incident hyperedges from $\mathcal{F}$, formally defined as follows: 
\[
\mathcal{F}_{e_s} = \{\,f^n\in \mathcal{F} : e_s \in f^n\}\,.
\]
Each hyperedge \(f^n\in \mathcal{F}_{e_s}\) defines an \(n\)-ary relation over a subset of $n$ entities.
To enable pairwise reasoning, we derive a set of pseudo-binary triples by enumerating ordered entity pairs within each hyperedge for query $q$ as follows:
\begin{equation}
\mathcal{T}_q = \{(e_h, f^n, e_t) \mid f^n\in \mathcal{F}_{e_s},\ e_h\in f^n, e_t \in f^n\},
\end{equation}
where each pseudo-binary triple \((e_h, f^n, e_t)\) consists of a head entity, the originating hyperedge, and a tail entity.

\subsubsection{Structural Proximity Encoding.}
To capture the structural proximity between entities in the hypergraph, we adapt the directional distance encoding (DDE) mechanism from SubGraphRAG~\cite{li2024subgraphrag}, extending it from binary relations to \(n\)-ary hyperedges.
Formally, for each candidate triple \((e_h, f^n,e_t) \in \mathcal{T}_q\), we compute its directional encoding in the following steps:
\begin{itemize}[leftmargin=*]
    \item \textbf{One‐Hot Initialization:} 
    For each entity \((e_h,f^n,e_t)\), we initialize a one-hot indicator for the head entity:
    \begin{equation}
    s_e^{(0)} =
    \left\{
    \begin{array}{ll}
    1, & \mbox{if } \exists (e_h, f^n, e_t) \in \mathcal{T}_q \mbox{ such that } e = e_h, \\
    0, & \mbox{otherwise}.
    \end{array}
    \right.
    \end{equation}
    \item \textbf{Bi-directional Feature Propagation:}
    For each layer \(l = 0, \dots, L\), we propagate features over the set of derived triples \(\mathcal{T}_q\).  
    Forward propagation simulates how the head entity \(e_h\) reaches out to the tail entity \(e_t\) as follows: 
    \begin{equation}
    s_e^{(l+1)} = \frac{1}{|\{e' \mid (e', \cdot, e) \in \mathcal{T}_q\}|} \sum_{(e', \cdot, e) \in \mathcal{T}_q} s_{e'}^{(l)}.
    \end{equation}
    In contrast, backward propagation updates head encodings based on tail-to-head influence:
    \begin{equation}
    s_e^{(r, l+1)} = \frac{1}{|\{e' \mid (e, \cdot, e') \in \mathcal{T}_q\}|} \sum_{(e, \cdot, e') \in \mathcal{T}_q} s_{e'}^{(r, l)}.
    \end{equation}
    \item \textbf{Bi-directional Encoding:}
    After \(L\) rounds of propagation, we concatenate the forward and backward encodings to obtain the final vector for each entity \(e\) as follows:
    \begin{equation}
      s_e
      = [\,s_e^{(0)} \,\|\, s_e^{(1)} \,\|\, \cdots \,\|\, s_e^{(L)}
        \,\|\, s_e^{(r,1)} \,\|\, \cdots \,\|\, s_e^{(r,L)}],
    \end{equation}
    where \(\|\) denotes vector concatenation. Note that the backward propagation starts from \(l=1\), as \(l=0\) is shared in both directions.
    \item \textbf{Triple Encoding:}
    For each candidate triple \((e_h,f^n,e_t)\), we define its structural proximity encoding as follows:
    \begin{equation}
    \delta(e_h, f^n, e_t) = \big[s_{e_h} \,\|\, s_{e_t}\big],
    \end{equation}
   which is passed to a lightweight parametric neural function to compute the plausibility score for each candidate triple \((e_h,f^n,e_t)\) given query $q$.
\end{itemize}

\subsubsection{Contrastive Plausibility Scoring.}
To reduce the search space in the hypergraph structure, we address the challenge that similarity-based retrieval often introduces noisy or irrelevant triples. To mitigate this, we train a lightweight MLP classifier \(f_{\theta}\) to estimate the plausibility of each triple candidate and prune uninformative ones.

To this end, the training set is prepared with positive and negative samples. 
Let \(P^*_q\) denote the shortest path of triples connecting the topic entity to a correct answer in the hypergraph \(\mathcal{G}\).
The positive samples \(\mathcal{T}^+_i\) at hop \(i\) consist of triples in \(P^*_q\), denoted as \(\mathcal{T}^+_i = \{(e_{h,i}, f^n_i, e_{t,i})\}\).
Negative samples \(T^-_i\) consist of all other triples incident to the head entity \(e_i\) at hop \(i\) that are not in \(P^*_q\).
At each exploration step, only positive triples are expanded at each hop, while negative ones are excluded. Each triple \((e_h,f^n,e_t)\) is encoded in a feature vector by concatenating its contextual and structural encodings:
\begin{equation}
\textbf{x} = \bigl[\varphi(q) \parallel \varphi(e_h) \parallel \varphi(f^n) \parallel \varphi(e_t) \parallel \delta(e_h,f^n,e_t)\bigr],
\end{equation}
where $\varphi$ denotes an embedding model that maps the textual content of the query ($q$), head entity ($e_h$), hyperedge ($f^n$), and tail entity ($e_t$), into vector representations, forming the candidate pseudo-binary triple $(e_h, f^n, e_t)$.
The classifier outputs a plausibility score \(f_{\theta}(\textbf{x}) \in [0,1]\), trained using binary cross-entropy as follows:
\begin{equation}
    \mathcal{L} = -\frac{1}{N}
  \sum_{i=1}^{N}
  \Bigl[
    y_i \,\log\bigl(f_{\theta}(\mathbf{x}_i)\bigr)
    \;+\;
    (1 - y_i)\,\log\bigl(1 - f_{\theta}(\mathbf{x}_i)\bigr)
  \Bigr].
\end{equation}

\subsubsection{Adaptive Search.}
At inference time, we initiate the retrieval process with initial triples of topic entities and compute their plausibility scores using the trained MLP, $f_{\theta}(\textbf{x})$.
Triples exceeding a plausibility threshold \(\tau\) are retained, and their tail entities are used as frontier entities in the next hop. This expansion–filtering cycle continues until no new triples satisfy the threshold.
However, using a fixed threshold \(\tau\) can be problematic: it may be too strict in sparse hypergraphs, limiting retrieval, or too lenient in dense hypergraphs, leading to an overload of irrelevant triples. To mitigate this, we implement an adaptive thresholding strategy. 
We initialize with \(\tau_0 = 0.5\), allow a maximum of \(N_{\max}=5\) threshold reductions, and define \(M=50\) as the minimum acceptable number of hyperedges per hop. At hop \(i\), we retrieve the set of triples, $\mathcal{T}_{q,\ge\tau_j}=\{(e_h,\mathbf{h},e_t)\mid f_\theta(x)\ge\tau_j\}$
under the current threshold \(\tau_j\). If \(|\mathcal{T}_{q,\ge\tau_j}| < M\), we iteratively reduce the threshold as follows: 
\begin{equation}
    \tau_{j+1} = \tau_{j} - c,\quad j = 0, \ldots, N_{\max} - 1,
\end{equation}
where $c=0.1$ is the decay factor. This process continues until \(\lvert |\mathcal{T}_{q,\ge\tau_j}|\rvert\ge M\) or the reduction limit is reached.
To further adapt to structural variations in the hypergraph, we incorporate a density-aware thresholding policy. Given the density of the hypergraph $\Delta(\mathcal{G})$ and the predefined lower and upper bounds $\Delta_\text{lo}$ and $\Delta_{\text{up}}$, we classify the hypergraph and adjust $\tau_0$ accordingly to balance coverage and precision as follows:
\begin{equation}
   \mathcal{M}_{\mathcal{G}} =
    \left\{
    \begin{array}{ll}
    \mathcal{M}_{\text{low}}, & \Delta(\mathcal{G})\le \Delta_{\text{lo}},\\
    \mathcal{M}_{\text{mid}}, & \Delta_{\text{lo}} < \Delta(\mathcal{G}) \le \Delta_{\text{up}}, \\
    \mathcal{M}_{\text{high}}, & \Delta(\mathcal{G}) > \Delta_{\text{up}}
    \end{array}
    \right.
\end{equation}
After convergence or exhaustion of threshold reduction attempts, the retrieval strategy is adjusted based on the assigned graph density category. For low-density graphs ($\mathcal{M}_{\text{low}}$), the retriever selects from previously discarded triples those that satisfy the final plausibility threshold. For medium and high-density graphs (\(\mathcal{M}_{\text{mid}}\) and \(\mathcal{M}_{\text{high}}\)), the strategy additionally expands from the tail entities of these newly accepted triples to increase the depth of reasoning.
This density-aware adjustment prevents over-retrieval in sparse graphs while enabling more profound and broader exploration in dense graphs. To further control expansion in high-density settings, where the number of candidate hyperedges may become excessive, we impose an upper bound on the number of retrieved triples per hop. This constraint effectively limits entity expansion, accelerates retrieval, and reduces the inclusion of low-utility information.

\subsubsection{Budget-aware Contextualized Generator.}
After completion of the retrieval process, we organize the selected elements into a structured input for the generator. Following the context layout protocol of HyperGraphRAG~\cite{hypergraphrag2025}, we include (i) entities and their associated descriptions, (ii) hyperedges along with their participating entities, and (iii) supporting source text chunks linked to each entity or hyperedge.
Due to input length constraints, we prioritize components based on their utility. As shown in the ablation study of HyperGraphRAG, \(n\)-ary relational facts (i.e., hyperedges) contribute the most to reasoning performance, followed by entities and then source text. We therefore allocate the token budget accordingly: 50\% for hyperedges, 30\% for entities, and 20\% for source chunks.
To further maximize informativeness, we order hyperedges and entities according to their plausibility scores $f_{\theta}(\cdot)$, with graph connectivity as a secondary criterion. The selected components are then sequentially filled in the order: hyperedges, entities, and source chunks. Components are filled in priority order and any unused budget is passed to the next category.
The contextualized evidence resulting \texttt{ context}, together with the original query $q$, is then passed to the LLM to generate the final answer \texttt{Answer} as:
\begin{equation}\label{eq:answer}
\texttt{Answer} \;:=\; \mathrm{LLM}(\texttt{Context}, q).
\end{equation}

\subsection{HyperMemory: Relational Chain Extraction}
To improve interpretability and context awareness in path retrieval, we avoid naive top-$k$ heuristics with LLM-guided scoring that leverages the model’s parametric memory to assess the salience of hyperedges and entities. This enables retrieval to be guided by contextual priors and query intent, facilitating more targeted and meaningful relational exploration.

\subsubsection{Memory-Guided Beam Retriever.}
Specifically, we design beam search with width \(w = 3\) and depth \(d = 3\), where \(w\) denotes the number of paths ranked in the top order retained at each iteration, and \(d\) specifies the maximum number of expansion steps.
Following the process of the \textit{Learnable Relational Chain Retriever}, we begin by identifying the set of topic entities $\mathcal{E}_q$ from the input query $q$ using an LLM-based entity extractor.
For each topic entity \(e_s \in \mathcal{E}_{q}\), we retrieve its incident hyperedge set \(\mathcal{F}_{e_{s}}\). Each hyperedge \(f^n \in \mathcal{F}_{e_{s}}\) is scored for relevance to both \(e_{s}\) and \(q\) using a prompt $p_{\text{edge}}$:
\begin{equation}
\mathcal{S}_{\mathcal{F}}(f^n \mid e_{s}, q) \sim \mathrm{LLM}(p_{\text{edge}}, e_{s}, f^n, q).
\end{equation}
We retain the top-\(w\) hyperedges, denoted \(H^{+}_{e_{s}}\), based on the score $\mathcal{S}_{\mathcal{F}}(\cdot)$. Next, for each \(f^n\in \mathcal{F}^{+}_{e_{s}}\), we identify unvisited tail entities \(e_{t}\) and score their relevance using a second prompt \(p_{\text{entity}}\):
\begin{equation}
\mathcal{S}_{\mathcal{E}}(e_{t}\mid f^n ,q) \;\sim \;\mathrm{LLM}(p_{\text{entity}},\ f^n, \ e_{t},\ q).
\end{equation}
Next, each resulting candidate triple \((e_{s},f^n,e_{t})\) receives a weighted composite score as follows:
\begin{equation}
\mathcal{S}(e_{s},f^n,e_{t}) = \mathcal{S}_{\mathcal{F}}(f^n \mid e_{s},q) \;\cdot\; \mathcal{S}_{\mathcal{E}}(e_{t}\mid f^n,q).
\end{equation}
From the current set of candidate triples, we retain the top-$w$ based on the final triple scorer \(\mathcal{S}(\cdot)\). The tail entities of these selected paths define the next expansion frontier.
At each depth $i$, we evaluate whether the accumulated evidence suffices to answer the query. All retrieved triples are assembled into a contextualized component \(C_{i}\), which is passed to the LLM for an evidence sufficiency check:
\begin{equation}
\mathrm{LLM}(p_{\text{ctx}},\,C_{i},\,q) \; \longrightarrow\; \{\texttt{yes},\texttt{no}\}, \texttt{Reason}.
\end{equation}
If the result is \texttt{yes}, terminate the search and proceed to generation. Otherwise, if \(i<d\), the search continues until the next iteration.

\subsubsection{Contextualized Generator.}
The entities and hyperedges retrieved are organized in a fixed format context, as defined in Eq.(\ref{eq:answer}). This contextualized evidence \texttt{Context}, combined with the original query $q$, is then passed to the LLM to generate the final \texttt{Answer}.

\begin{table*}[t]
\centering
\setlength\tabcolsep{1pt}
\resizebox{0.85\textwidth}{!}{%
\begin{tabular}{l r@{\hskip 4pt}c
                r@{\hskip 4pt}c
                r@{\hskip 4pt}c
                r@{\hskip 4pt}c|
                r@{\hskip 4pt}c
                r@{\hskip 4pt}c
                r@{\hskip 4pt}r}
\toprule
\multirow{2}{*}{\textbf{Topic}} 
  & \multicolumn{2}{c}{\textbf{RAPTOR}}
  & \multicolumn{2}{c}{\textbf{HippoRAG}} 
  & \multicolumn{2}{c}{\textbf{ToG}} 
  & \multicolumn{2}{c}{\textbf{HyperGraphRAG}}
  & \multicolumn{2}{|c}{\textbf{HyperRetriever}} 
  & \multicolumn{2}{c}{\textbf{HyperMemory}} 
  & \multicolumn{2}{c}{\textbf{Rel. Gain (\%)}} \\
\cmidrule(r){2-3} \cmidrule(lr){4-5} \cmidrule(lr){6-7} 
\cmidrule(lr){8-9} \cmidrule{10-11} \cmidrule(l){12-13} \cmidrule(l){14-15}
& MRR & Hits@10  
& MRR & Hits@10  
& MRR & Hits@10  
& MRR & Hits@10 
& MRR & Hits@10  
& MRR & Hits@10  
& MRR & Hits@10  \\
\midrule
\textsc{art}     &  3.44  &  4.13 &  8.42  &  9.77  &  2.99  &  3.20  &  \underline{17.18} &  \underline{21.68} &  \textbf{19.31} &  \textbf{24.31}  &  15.63 &  19.17  & 12.40 & 12.13 \\
\textsc{award}   &  20.57  &  25.13 &  32.80  &  38.65  &  8.70  &  9.35  &  \underline{51.64} &  \underline{63.43} &  \textbf{52.66} &  \textbf{65.28}  &  47.34 &  56.98  & 1.98 & 2.93 \\
\textsc{edu}     &  4.94  &  5.90 &  23.82  &  26.37  &  9.09  &  9.49  &  \underline{43.44} &  \underline{50.05} &  \textbf{44.79} &  \textbf{51.63}  &  41.68 &  46.95  & 3.11 & 3.16 \\
\textsc{health}  &  18.85  &  22.04 &  25.72  &  29.59  &  7.14  &  7.95  &  \underline{31.46} &  \underline{37.94} &  \textbf{32.68}  &  \textbf{39.26}  &  27.48 &  33.13  &  3.88  & 3.48 \\
\textsc{infra}   &  10.95 &  12.79 &  23.88  &  27.11  &  9.87  &  10.67 &  \underline{37.18} &  \underline{44.82} &  \textbf{38.92} &  \textbf{45.77}  &  35.77 &  41.69  & 4.68 & 2.12 \\
\textsc{loc}     &  16.55 &  18.68 &  19.88 &  23.08 &  3.45  &  3.83  &  29.92 &  34.38 &  \textbf{31.80} &  \textbf{36.85}  &  \underline{30.73} &  \underline{35.95}  & 6.28 & 7.18 \\
\textsc{org}     &  12.00  &  14.54 &  36.20  &  41.70  &  6.61  &  7.33  &  \textbf{64.68} &  \textbf{74.89} &  \underline{62.87} &  \underline{71.21}  &  52.26 &  59.84  & -2.80 & -4.91 \\
\textsc{people}  &  10.74  &  13.10  &  15.39  &  18.28  &  3.90  &  4.40  &  \underline{20.67} &  \underline{28.10} &  \textbf{21.62} &  \textbf{28.48}  &  18.96 &  25.29  & 4.60 & 1.35 \\
\textsc{sci}     &  6.84  &  8.66  &  15.62  &  18.86  &  6.87  &  7.28  &  \textbf{25.92} &  \textbf{34.54} &  \underline{25.15} &  \underline{32.30}  &  21.50 &  27.53  & -2.97 & -6.49 \\
\textsc{sport}   &  11.31 &  13.28 &  22.78 &  26.01 &  7.51  &  8.53  &  \underline{37.40} &  \underline{44.91} &  \textbf{39.37} &  \textbf{45.56}  &  33.64 &  39.72  & 5.27 & 1.45 \\
\textsc{tax}     &  10.48  &  11.08 &  24.77  &  26.65  &  6.22  &  6.50  &  \underline{35.15} &  \underline{40.94} &  \textbf{37.20} &  \textbf{40.98}  &  33.65 &  38.19  & 5.83 & 0.10 \\
\midrule
\textsc{AVG}     &  11.52 &  13.58 &  22.66  &  26.01  & 6.58  &  7.14  &  \underline{35.88}  &  \underline{43.24}  &  \textbf{36.94}  &  \textbf{43.78}  & 32.60 &  38.59 &  2.95  &  1.23 \\
\bottomrule
\end{tabular}%
}
\caption{Performance comparison of domain generalization across 11 diverse topics. 
The ``Rel. Gain'' column highlights the substantial relative improvement of our approach over the best baseline, averaged across all domains (metrics in \%).}
\label{table2}
\end{table*}

\begin{table}[t]
\centering
\setlength\tabcolsep{1pt}
\resizebox{0.96\linewidth}{!}{
\begin{tabular}{l *{6}{c}}
\toprule
\multirow{2.5}{*}{Model} & \multicolumn{2}{c}{HotpotQA} 
& \multicolumn{2}{c}{MuSiQue} 
& \multicolumn{2}{c}{2WikiMultiHopQA} \\
\cmidrule(lr){2-3} \cmidrule(lr){4-5} \cmidrule(lr){6-7}
 & EM(\%) & F1(\%)
 & EM(\%)  & F1(\%)
 & EM(\%)  & F1(\%) \\
\midrule
RAPTOR        & 35.50 & 41.56 & \underline{15.00} & 16.31 & 22.50 & 22.95 \\
HippoRAG      & \underline{49.50} & \textbf{55.87} & 14.50 & \underline{17.43} & 30.00 & 30.44 \\
ToG           & 10.08 & 11.00  & \phantom{0}2.70  & \phantom{0}2.69  & \phantom{0}5.20 & \phantom{0}5.34  \\
HyperGraphRAG & \textbf{51.00} & 42.69 & \textbf{22.00} & \textbf{20.02} & \textbf{42.50} & 30.17 \\
\midrule
HyperRetriever     & 42.50 & \underline{43.65} & 13.50 & 14.15 & \underline{34.00} & \textbf{34.06} \\
HyperMemory  & 35.50 & 41.51 & \phantom{0}8.00  & 12.96 & 31.50 & \underline{32.56} \\
\midrule
Rel. Gain (\%) &  -16.67 & -21.87 & -38.64 & -29.32 & -20.00 & 11.89  \\
\bottomrule
\end{tabular}
}
\caption{Performance comparison on HotpotQA, MuSiQue, and 2WikiMultiHopQA. Rel. Gain (\%) indicates the relative performance gains achieved by our model compared with the best baselines. The best results are \textbf{bolded}, and the second best are \underline{underlined}.}
\label{table1}
\end{table}


\section{Experiments}\label{sec:exp}
We quantitatively evaluate the effectiveness and efficiency of HyperRetriever against RAG baselines both in-domain and cross-domain settings. Ablation studies highlight the benefits of adaptive expansion and \(n\)-ary relational chain learning, complemented by qualitative analyzes that illustrate the precision and efficiency of the adaptive retrieval process.

\subsection{Experimental Setup}

\subsubsection{Datasets.}
We conduct experiments under both open-domain and closed-domain multi-hop question answering (QA) settings. For in-domain evaluation, we use three widely adopted benchmark datasets: HotpotQA \cite{hotpotqa2018}, MuSiQue \cite{musique2022}, and 2WikiMultiHopQA \cite{ho2020}. To evaluate cross-domain generalization, we adopt the WikiTopics-CLQA dataset \cite{gao2023double}, which tests zero-shot inductive reasoning over unseen entities and relations at inference time. 
Comprehensive dataset statistics are summarized in Appendix \ref{ap:dataset}.

\subsubsection{Evaluation Metrics.}
We employ four standard metrics to assess performance, aligning with established protocols for each benchmark type. For open-domain QA datasets, where the objective is precise answer generation, we report Exact Match (EM) and F1 scores. For WikiTopics-CLQA, which involves ranking correct entities from a candidate list, we utilize Mean Reciprocal Rank (MRR) and Hits@k to evaluate retrieval fidelity. All metrics are reported as percentages (\%), with higher values indicating better performance.

\subsubsection{Baselines.}
To evaluate the effectiveness of our approach, we compare HyperRAG with RAG baselines with varying retrieval granularities, enabling a systematic analysis of how evidence structure affects retrieval effectiveness and answer generation in both open- and closed-domain settings. Specifically, we include: RAPTOR \cite{sarthi2024raptor}, which retrieves tree-structured nodes; HippoRAG \cite{hipporag2024}, which retrieves free-text chunks; ToG \cite{sun2024thinkongraph}, which retrieves relational subgraphs; and HyperGraphRAG \cite{hypergraphrag2025}, which retrieves a heterogeneous mixture of entities, relations, and textual spans. 

\subsubsection{Implementation Details.}
All baselines and our proposed methods utilize \texttt{gpt-4o-mini} as the core model for both graph construction and question answering. For HyperRetriever, we additionally employ the pretrained text encoder \texttt{gte-large-en-v1.5} to produce dense embeddings for entities, relations, and queries. With 434M parameters, this GTE-family model achieves strong performance on English retrieval benchmarks, such as MTEB, and offers an efficient balance between inference speed and embedding quality, making it well-suited for semantic subgraph retrieval. 
All experiments were implemented in Python 3.11.13 with CUDA 12.8 and conducted on a single NVIDIA RTX 3090 (24 GB). Peak GPU memory usage remained within 24 GB due to dynamic allocation.

\begin{table*}[h]
\centering
\setlength\tabcolsep{2pt}
\resizebox{0.8\textwidth}{!}{%
\begin{tabular}{l r@{\hskip 4pt}c
                r@{\hskip 4pt}c
                r@{\hskip 4pt}c
                r@{\hskip 4pt}c
                r@{\hskip 4pt}c
                r@{\hskip 4pt}c
                r@{\hskip 4pt}c
                r@{\hskip 4pt}r}
\toprule
\multirow{2}{*}{\textbf{Topic}} 
  & \multicolumn{2}{c}{\textbf{Full}}
  & \multicolumn{2}{c}{\textbf{w/o Entities}} 
  & \multicolumn{2}{c}{\textbf{w/o Hyperedges}} 
  & \multicolumn{2}{c}{\textbf{w/o Chunks}}
  & \multicolumn{2}{c}{\textbf{w/o Adaptive Search}} 
  & \multicolumn{2}{c}{\textbf{w Binary KG}} \\
\cmidrule(r){2-3} \cmidrule(lr){4-5} \cmidrule(lr){6-7} 
\cmidrule(lr){8-9} \cmidrule{10-11} \cmidrule{12-13}
& MRR & Hits@10  
& MRR & Hits@10  
& MRR & Hits@10  
& MRR & Hits@10 
& MRR & Hits@10  
& MRR & Hits@10 \\
\midrule
\textsc{art}     & 26.03 & \underline{31.00} & \textbf{27.28} & 31.00 & 24.03 & 27.00 & 24.17 & 27.00 & 26.33 & \underline{31.00} & 14.00 & 15.00 \\
\textsc{award}   & \textbf{56.91} & \underline{70.00} & 43.22 & 61.00 & 55.95 & 69.00 & 55.01 & 66.00 & 52.98 & 66.00 & 48.92 & 53.00 \\
\textsc{edu}     & \textbf{49.00} & \underline{56.00} & 43.24 & 52.00 & 47.93 & 52.00 & 42.67 & 47.00 & 47.53 & 53.00 & 38.20 & 42.00 \\
\textsc{health}  & \textbf{41.25} & \underline{47.00} & 37.17 & 43.00 & 37.70 & 40.00 & 39.33 & \underline{47.00} & 39.20 & 46.00 & 36.17 & 39.00 \\
\textsc{infra}   & 34.85 & 43.00 & 35.17 & 43.00 & 30.87 & 39.00 & \textbf{38.75} & 44.00 & 35.50 & \underline{45.00} & 30.50 & 32.00 \\
\textsc{loc}     & 38.75 & 42.50 & \textbf{44.58} & \underline{47.50} & 37.50 & 40.00 & 33.13 & 37.50 & 41.67 & \underline{47.50} & 39.58 & 42.50 \\
\textsc{org}     & 46.79 & 58.97 & \textbf{58.75} & \underline{65.00} & 45.92 & 55.00 & 53.00 & 60.00 & 38.07 & 45.00 & 47.50 & 47.50 \\
\textsc{people}  & 14.20 & 22.00 & \textbf{21.23} & \underline{28.00} & 13.73 & 19.00 & 20.03 & 26.00 & 13.37 & 20.00 & 19.33 & 22.00 \\
\textsc{sci}     & 25.91 & 36.00 & 18.67 & 22.00 & 24.53 & 32.00 & \textbf{26.09} & \underline{38.00} & 21.14 & 32.00 & 24.00 & 27.00 \\
\textsc{sport}   & 31.04 & 40.00 & 35.83 & 40.00 & 35.00 & 45.50 & 29.58 & 40.00 & 33.33 & 37.50 & \textbf{42.08} & \underline{47.50} \\
\textsc{tax}     & 36.25 & \underline{40.00} & 29.17 & 35.00 & 33.54 & 36.25 & 33.13 & 36.25 & \textbf{36.88} & \underline{40.00} & 35.42 & 37.50 \\
\midrule
\textsc{AVG}     & \textbf{36.45} & 40.59 & 35.85 & 42.50 & 35.15 & 41.34 & 35.90 & 42.61 & 35.64 & \underline{42.91} & 34.15 & 36.82 \\
\bottomrule
\end{tabular}%
}
\caption{Ablation on the Contribution of Context Formation and Adaptive Search.
The full model incorporates all components essential for context formation, including entities, hyperedges involved in learnable relational chains, and retrieved chunks. The best results in MRR are \textbf{bolded}, and the best in Hits@10 are \underline{underlined}.}
\label{table3}
\end{table*}

\subsection{Open-domain Answering Performance}\label{sec:rq1} 
\subsubsection{Setup}
For \textbf{HyperRetriever}, a lightweight MLP $f_\theta$ scores the plausibility of candidate hyperedges, enabling aggressive pruning that reduces traversal complexity without compromising reasoning quality.
For \textbf{HyperMemory}, we set beam width $w=3$ and depth $d=3$ to balance retrieval coverage against computational cost.
Comprehensive prompt definitions for edge scoring ($p_{\text{edge}}$), entity ranking ($p_{\text{entity}}$), context evaluation ($p_{\text{ctx}}$), and generation are provided in the Appendix.

\subsubsection{Results}
Table~\ref{table1} details the Exact Match (EM) and F1 scores across three open-domain QA benchmarks. HyperRetriever consistently outperforms the HyperMemory variant on HotpotQA and MuSiQue, demonstrating superior capability in identifying evidential relational chains. This advantage is attributed to its learnable MLP-based plausibility scorer and density-aware expansion strategy, which affords precise control over retrieval depth. In contrast, HyperMemory relies on the fixed parametric memory of the LLM, rendering it less adaptable to domain-specific relational patterns.
When compared to external KG-based RAG baselines, we observe a performance divergence based on graph topology. On HotpotQA and MuSiQue, HyperRetriever exhibits a performance gap (e.g., 38.64\% lower EM on MuSiQue), likely because these datasets require the rigid structural guidance of explicit KG priors for cross-document navigation. However, on 2WikiMultiHopQA, HyperRetriever reverses this trend, achieving an 11.89\% relative F1 improvement. This suggests that while KG priors aid in sparse settings, HyperRetriever is uniquely effective at exploiting the denser, complex relational contexts found in 2WikiMultiHopQA.


\subsection{Closed-domain Generalization Performance}\label{sec:rq2}

To evaluate adaptability to closed-domain \(n\)-ary knowledge graphs, we evaluate the performance of \textbf{HyperRAG} on the WikiTopics-CLQA dataset (Table~\ref{table2}). The results demonstrate a strong generalization across diverse topic-specific hypergraphs. In particular, our learnable variant, HyperRetriever, achieved the highest overall answer precision, with average improvements of 2.95\% (MRR) and 1.23\% (Hits@10) compared to the second-best baseline, HyperGraphRAG. 
These gains are statistically significant ($p \ll 0.001$), with $t$-test values of $1.46\times10^{-17}$ for MRR and $2.41\times10^{-6}$ for Hits@10, suggesting the empirical reliability of our approach.
HyperRetriever secures top performance in 9 out of the 11 categories—for instance, achieving relative gains of 12.40\% (MRR) and 12.13\% (Hits@10) in the \textsc{Art} domain—and consistently ranks second in the remaining two. 
This broad efficacy highlights the robustness of HyperRetriever's adaptive retrieval mechanism. 
Unlike baselines that are sensitive to domain-specific graph density, HyperRetriever’s learnable MLP scorer dynamically calibrates its expansion strategy to suit varying $n$-ary topologies, ensuring high precision even in complex reasoning tasks.
In contrast, our memory-guided variant, \textbf{HyperMemory}, consistently underperforms against to HyperRetriever. 
This variant serves as a critical ablation to probe the limitations of an LLM’s intrinsic parametric memory for $n$-ary retrieval. 
The results confirm that prompt-based scoring alone, without the explicit structural learning provided by HyperRetriever, is insufficient for multi-hop reasoning in closed domains.

\begin{table*}[t]
\centering
\setlength\tabcolsep{1.5pt}
\renewcommand{\arraystretch}{1.12}
\resizebox{\textwidth}{!}{
\begin{tabular}{lcccccc}
\toprule
\textbf{Dimension} &
\textbf{RAPTOR \cite{sarthi2024raptor}} &
\textbf{HippoRAG \cite{hipporag2024}} &
\textbf{ToG \cite{sun2024thinkongraph}} &
\textbf{HyperGraphRAG \cite{hypergraphrag2025}} &
\textbf{OG-RAG \cite{ograg2024}} &
\textbf{HyperRetriever / Memory} \\
\midrule
Structure type                    & Doc tree (summ.)  & KG (binary)           & KG (binary)        & Hypergraph (\(n\)-ary) & Object graph (mostly bin.) & Hypergraph (\(n\)-ary) \\
Unit of fact                 & Passage / summary & Entity-entity edge    & Step / subgoal       & Hyperedge (\(n\)-ary fact)   & Object-object edge & Hyperedge (\(n\)-ary fact) \\
Candidate growth   & Additive (levels) & Additive on edge & LLM-var.  & Additive on hyperedges    & Additive on objects & Additive on hyperedges \\
Per-query overhead   & Tokens only       & \(\mathcal{O}(n{-}k)\) & Var.           & \(\mathcal{O}(1)\)\textsuperscript{\dag} & \(\mathcal{O}(1)\) & \(\mathcal{O}(1)\)\textsuperscript{\dag} \\
Depth for reasoning chain & Deep              & Deep (pairwise)      & LLM-var.         & Shallow (\(n\)-ary edges)     & Deep (pairwise)   & Shallow (\(n\)-ary edges) \\
Retrieval strategy               & Dense tree search & Graph walk + dense     & LLM on graph     & Static    & Object-centric walk & Adaptive / LLM on graph\\
LLM at retrieval             & Low-Med          & Low                    & Med-High (LLM)  & Low                       & Low                 & Low / Med (LLM) \\
Ontology                     & \xmark          & \xmark                     & \xmark   & \xmark                       & \cmark                & \xmark  \\
\bottomrule
\end{tabular}}
\caption{Method Comparison. 
HyperRetriever utilizes adaptive search on \(n\)ary hyperedges, enabling higher-order reasoning with shallow chains and near constant per-query retrieval overhead \(\mathcal{O}(1)\). 
In contrast, static or object-centric walks on binary graphs entail deeper pairwise chains and materialization cost.
$\dagger$ denotes bounded arity; \cmark\ indicates an ontology requirement.}
\label{tab:related_work}
\end{table*}

\subsection{Ablation Study}\label{sec:rq3}
To evaluate the effectiveness of our approach, we conduct a series of ablation studies targeting two key aspects: (i) the contribution of individual components to context formation, and (ii) the impact of the adaptive search policy on retrieval performance.

\subsubsection{Higher-Order Reasoning Chains}
Compared with binary KG RAG, \emph{HyperRAG} supports higher-order reasoning on 
\(n\)-ary hypergraphs. An \(n\)-ary hyperedge jointly binds multiple entities and roles, capturing fine-grained dependencies beyond pairwise links. Exploiting this structure yields shallower yet more expressive reasoning chains, enabling the model to surface key evidence without multi-hop traversal. Empirically (Table~3), replacing the \(n\)-ary structure with a binary one lowers average MRR from \(36.45\%\) to \(34.15\%\) (-2.3\%) and the average Hits @ 10 from \(40.59\%\) to \(36.82\%\) (-3.77\%), indicating gains in both accuracy and efficiency. Additional qualitative examples appear in Appendix~\ref{ap:quality}.

\subsubsection{Impact of Context Formation}
Table~\ref{table3} presents a component-wise ablation study conducted on a representative 1\% subset to isolate the contributions of (i) entities, (ii) structural relations (hyperedges), and (iii) textual context. We observe that removing any component consistently degrades Mean Reciprocal Rank (MRR), though Hits@10 exhibits higher variance. This divergence highlights the distinction between ranking fidelity (MRR) and candidate inclusion (Hits@10). For instance, in the \textsc{org} and \textsc{loc} domains, certain ablated variants maintain competitive Hits@10 scores but suffer sharp declines in MRR. This indicates that while the correct answer remains within the top candidates, the loss of structural or semantic signals causes it to drift down the ranking list, degrading precision.
Crucially, hyperedges emerge as the dominant factor in effective context formation. Their exclusion precipitates the most significant performance drops across both metrics, underscoring the necessity of high-order topological structure for reasoning. In contrast, removing entities yields less severe degradation, as entities primarily provide node-level descriptions, whereas hyperedges capture the joint dependencies between them. Text chunks offer complementary unstructured semantics but lack the relational precision of the graph structure. Ultimately, the superior performance of the full model validates the synergistic integration of entity-aware signals, hypergraph topology, and adaptive textual evidence.



\subsubsection{Impact of Adaptive Search}
Removing the adaptive search component results in a noticeable decline in MRR across most categories, whereas its impact on Hit@10 is minimal and in some cases (e.g., \textsc{infra}, \textsc{loc}), even marginally positive. This pattern suggests that while correct answers remain retrievable among the top 10 candidates, they tend to be ranked lower in the absence of adaptive search, resulting in a reduced overall ranking precision. 

\subsection{Efficiency Study}\label{sec:rq4}
\subsubsection{Setup} 
\begin{figure}[t]
    \centering
    \captionsetup{type=figure}
    \includegraphics[width=0.82\linewidth]{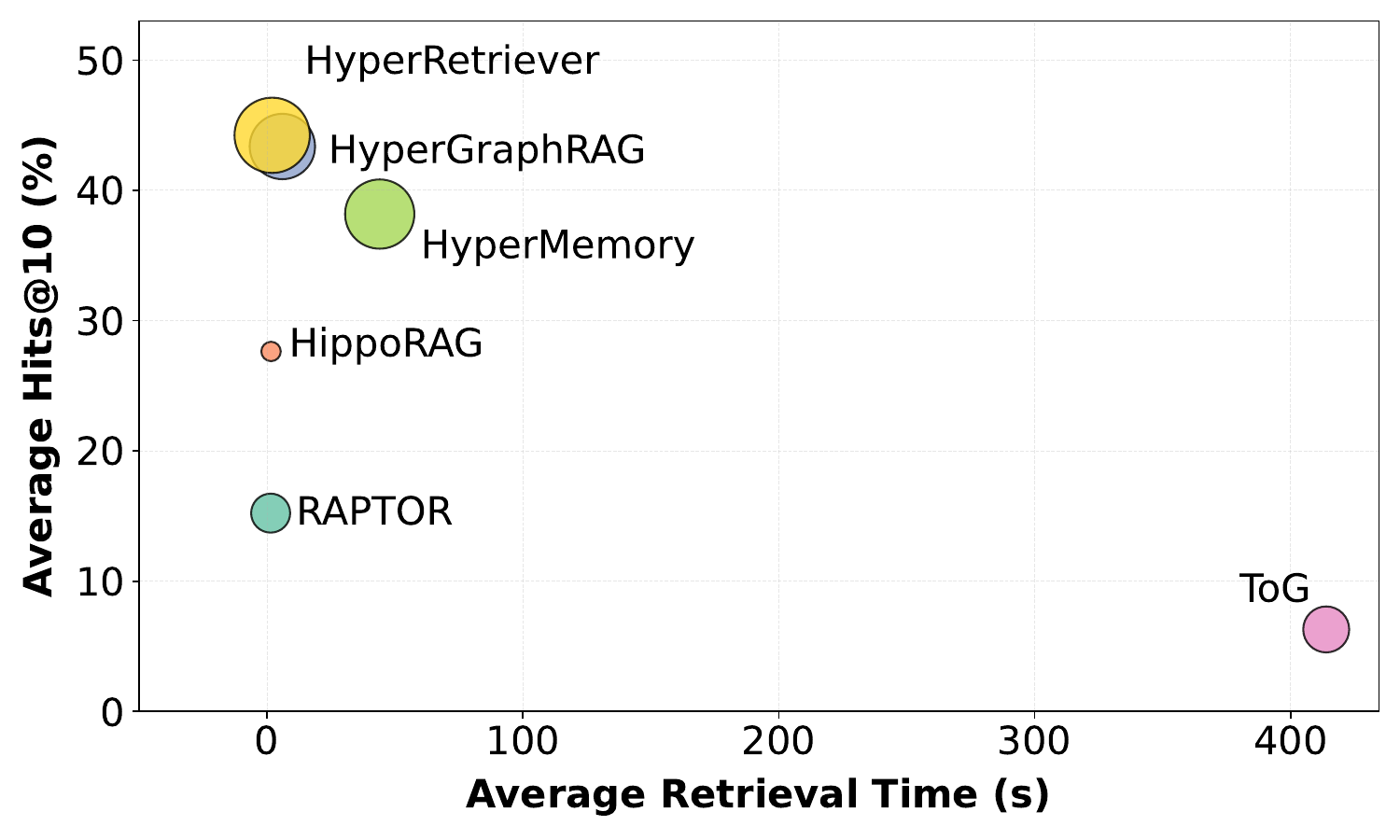}
    \captionof{figure}{
    The visualization shows the efficiency-effectiveness tradeoff in multi-hop QA: retrieval time ($x$-axis), answer quality (Hits@10, $y$-axis), and context volume (bubble size, log-scaled by retrieved tokens).}\label{fig:cp}
\end{figure}

To assess retrieval efficiency, we draw a stratified 1\% from each WikiTopics-CLQA category, yielding approximately 1,000 questions evenly distributed across 11 topic domains, and evaluate all baselines on this set. Figure~\ref{fig:cp} depicts the three-way trade off among retrieval time ($x$-axis), Hits@10 accuracy ($y$-axis), and context volume (bubble size, logarithmically scaled by retrieved tokens). Models in the upper left quadrant achieve the best balance between efficiency and effectiveness, combining low latency with high Hits@10 while retrieving compact contexts.

\subsubsection{Empirical Evidence} 
HyperRetriever achieves the shortest retrieval time and the highest Hits@10. Although it retrieves more tokens than some baselines, top performers consistently rely on larger contexts, highlighting a common trade-off between answer quality and retrieval volume. 
Our empirical findings align with the theoretical analysis in \S\ref{sec:problem}. \emph{HyperRetriever} employs adaptive search over \(n\)-ary hyperedges, enabling higher-order reasoning with shallow chains and nearly constant per query overhead \(\mathcal{O}(1)\).
In contrast, static or object-centric walks in binary graphs require deeper pairwise chains and incur an event materialization cost \(\mathcal{O}(n{-}k)\).
We further benchmark our approach against five publicly available graph-based RAG systems, covering both \(n\)-ary and binary KG designs, and summarize in Table~\ref{tab:related_work}.



\section{Related Work}\label{sec:work}

\noindent \textbf{Retrieval-Augmented Generation. }
RAG fundamentally augments the parametric memory of LLMs with external data, serving as a critical countermeasure against hallucination in knowledge-intensive tasks. 
The standard pipeline operates by retrieving top-$k$ document chunks via dense similarity search before conditioning generation on this augmented context~\cite{angeli2015,gao2024,Huang_2025}. 
However, conventional dense retrieval methods~\cite{moreira2024,lee2025nvembed} treat data as flat text, often overlooking the complex structural and relational signals required for deep reasoning. 
To address this, iterative multi-step retrieval approaches have been proposed~\cite{su-etal-2024-dragin,jiang-etal-2023-active,trivedi-etal-2023-interleaving}. 
Yet, these methods often suffer from diminishing returns: they increase inference latency and retrieve redundant information that dilutes the context signal. 
This noise contributes to the ``lost-in-the-middle'' effect, where finite context windows prevent the LLM from effectively attending to dispersed evidence~\cite{xu2024retrieval,liu-etal-2024-lost}.


\noindent \textbf{Graph-based RAG. }
Graph-based RAG frameworks incorporate inter-document and inter-entity relationships into retrieval to enhance coverage and contextual relevance \cite{10.5555/2846367, peng2024graphrag, han2025graphrag,barry-etal-2025}. Early approaches queried curated KGs (e.g., WikiData, Freebase) for factual triples or reasoning chains \cite{10.1145/2629489, 10.1145/1376616.1376746, luo2024rog, li-etal-2023-graph}, while recent methods fuse KGs with unstructured text \cite{edge2025localglobalgraphrag, 10.1145/3701716.3715240} or build task-specific graphs from raw corpora \cite{abs-2405-18414}.
To improve efficiency, LightRAG \cite{guo2024lightrag}, HippoRAG \cite{hipporag2024}, and MiniRAG \cite{fan2025minirag} adopt graph indexing via entity links, personalized PageRank, or incremental updates \cite{luo2025gfmrag, gnnkarypis-2025}. However, KG-based RAGs often face a trade-off between breadth and precision: broader retrieval increases noise, while narrower retrieval risks omitting key evidence. Methods using fixed substructures (e.g., paths, chunks) simplify reasoning \cite{sarthi2024raptor, graphbasedrag2025} but may miss global context, and challenges are amplified by LLM context window limits, vast KG search spaces \cite{jiang-etal-2023-active, Pan_2024, sun2024thinkongraph}, and the high latency of iterative queries \cite{sun2024thinkongraph}.
Moreover, most graph-based RAG methods rely on binary relational facts, limiting the expressiveness and coverage of knowledge. Hypergraph-based representations capture richer \(n\)-ary relational structures \cite{text2nkg2024}. HyperGraphRAG \cite{hypergraphrag2025} advances this line by leveraging \(n\)-ary hypergraphs, outperforming conventional KG-based RAGs, yet suffers from noisy retrieval and reliance on dense retrievers. OG-RAG \cite{ograg2024} addresses these issues by grounding hyperedge construction and retrieval in domain-specific ontologies, enabling more accurate and interpretable evidence aggregation. However, its dependence on high-quality ontologies constrains scalability in fast-changing or low-resource domains. Most graph-based and hypergraph-based RAG methods still face challenges, particularly due to the use of static or object-centric walks on binary graphs, which entail deeper pairwise chains and higher materialization costs. Table~\ref{tab:related_work} compares existing methods with \emph{HyperRAG}.


\section{Conclusion}
We introduced HyperRAG, a novel framework that advances multi-hop Question Answering by shifting the retrieval paradigm from binary triples to $n$-ary hypergraphs featuring two strategies: HyperRetriever, designed for precise, structure-aware evidential reasoning, and HyperMemory, which leverages dynamic, memory-guided path expansion. Empirical results demonstrate that HyperRAG effectively bridges reasoning gaps by enabling shallower, more semantically complete retrieval chains. Notably, HyperRetriever consistently outperforms strong baselines across diverse open- and closed-domain datasets, proving that modeling high-order dependencies is crucial for accurate and interpretable RAG systems.

\begin{acks}
This work is partially supported by the National Science and Technology Council (NSTC), Taiwan (Grants: NSTC-112-2221-E-A49-059-MY3, NSTC-112-2221-E-A49-094-MY3, 114-2222-E-A49-004, and 114-2639-E-A49-001-ASP).
\end{acks}

\bibliographystyle{ACM-Reference-Format}
\balance
\bibliography{sample-base}

\appendix
\clearpage
\section{Reduction to Binary Knowledge Graphs}\label{ap:reduction}

\begin{definition}[Faithful Reduction to Binaries]
Let $\mathcal{F}$ be a set of $n$-ary facts $(n\!\ge\!3)$ over entities $\mathcal{E}$ with role-typed arguments. A \emph{reduction} is a mapping $\Phi:\mathcal{P}(\mathcal{F})\!\to\!\mathcal{P}(\mathcal{E}\!\times\!\mathcal{E})$ that introduces no new auxiliary nodes and satisfies, for all $F,F'\!\subseteq\!\mathcal{F}$:
\begin{enumerate}[leftmargin=*]
    \item \textbf{Recoverability:} $F$ is uniquely determined by $\Phi(F)$ without spurious or missing tuples; and
    \item \textbf{Role preservation:} argument roles in $F$ are recoverable from $\Phi(F)$; and
    \item \textbf{Multiplicity:} distinct co-participation instances remain distinguishable (no accidental merging).
\end{enumerate}
\end{definition}
The three conditions cannot be met under a binary-only schema $\Phi$. Intuitively, triadic and higher-arity facts impose \emph{joint} constraints across all arguments, whereas binaries encode only \emph{pairwise} co-occurrence. Removing the joint carrier hyperedge either obscures “\textit{who did what with which role}” or merges parallel events. Therefore, an auxiliary event node (or equivalent mechanism) is necessary to preserve tuple identity and roles. An illustrative example follows.
\begin{example}[Role Ambiguity]
\[
\begin{aligned}
F_1 &= \{\textsf{give}(\text{Alice},\text{Bob},\text{Book}),\
       \textsf{give}(\text{Alice},\text{Carol},\text{Pen})\},\\
F_2 &= \{\textsf{give}(\text{Alice},\text{Bob},\text{Pen}),\
       \textsf{give}(\text{Alice},\text{Carol},\text{Book})\}.
\end{aligned}
\]
Naive pairwise projection (no event node):
\[
\Phi(F)=\left\{
\begin{aligned}
&\textsf{gaveTo}(\text{Alice},\text{Bob}),\ \textsf{gaveTo}(\text{Alice},\text{Carol}),\\
&\textsf{gaveItem}(\text{Alice},\text{Book}),\ \textsf{gaveItem}(\text{Alice},\text{Pen})
\end{aligned}
\right\}.
\]
Then $\Phi(F_1)=\Phi(F_2)$: the (receiver, item) pairing is unrecoverable, violating
\emph{recoverability} and \emph{role preservation}.
\end{example}

\begin{figure}[!ht]
    \centering
    \includegraphics[width=\columnwidth]{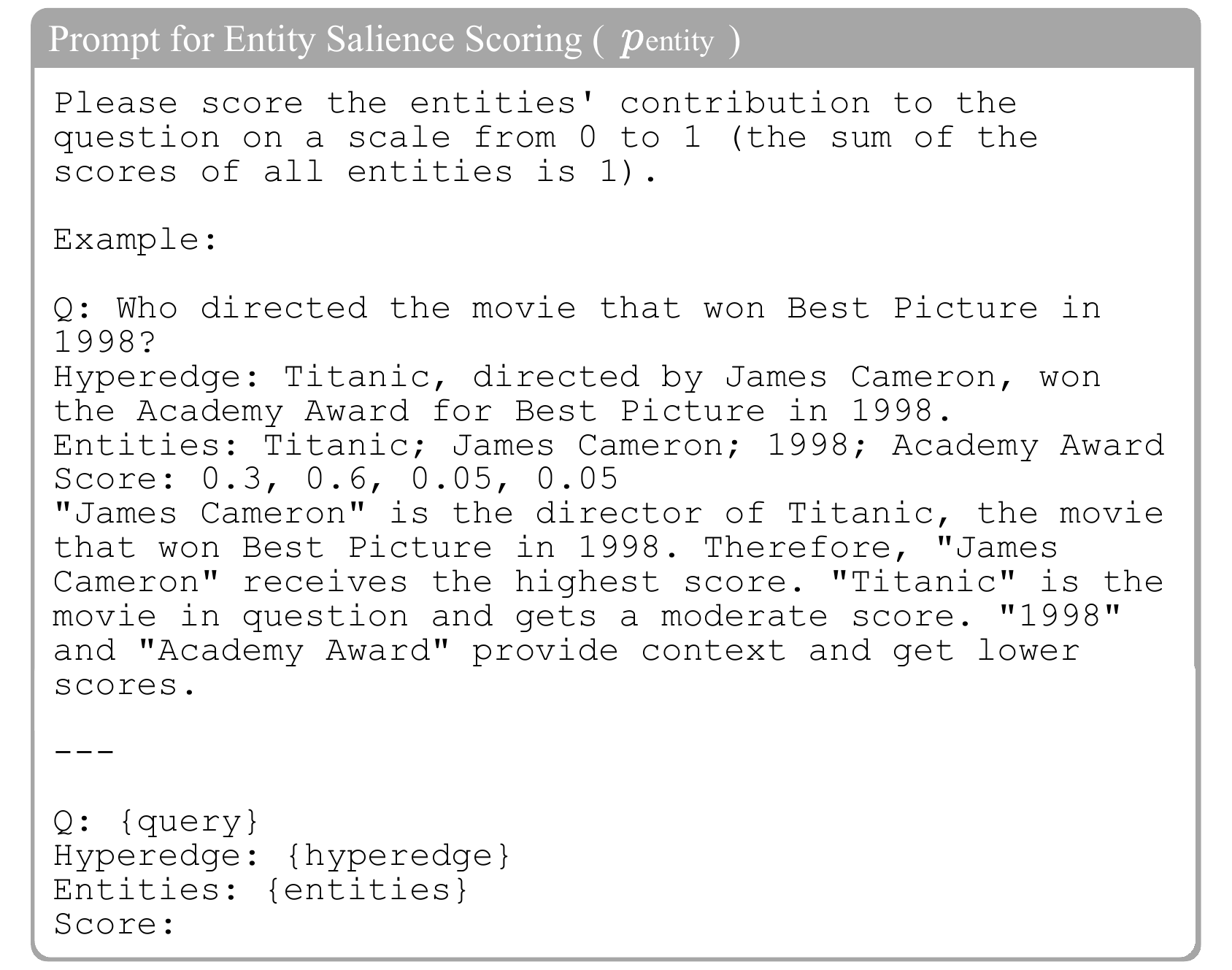}
    \caption{Prompt for Entity Salience Scoring ($p_{\text{entity}}$).}
    \label{fig:entity}
\end{figure}

\section{Reproducibility Details}\label{ap:hyperparameters}

\subsection{Hyperparameter Setting}\label{ap:hyperparameters}

\par\noindent \textbf{HyperRetriever Hyperparameters. }
HyperRetriever is trained using \texttt{nn.BCEWithLogitsLoss} with a batch size of 32, learning rate of $1\times10^{-4}$, and early stopping (patience = 10) over 50 epochs. 
For the retrieval phase of HyperRetriever, we followed the hyperparameters specified in the methodology: initial plausibility threshold $\tau_0=0.5$, maximum threshold reductions $N_{\max}=5$, minimum number of hyperedges per question $M=50$, and decay coefficient $c=0.1$. To further adapt retrieval behavior based on the graph structure, we design hypergraph's density lower and upper bounds $\Delta_{\text{lo}}=2.35$ and $Delta_{\text{up}}=5$.

\par\noindent \textbf{HyperMemory Hyperparameters. }
For HyperMemory, we set the beam width $w=3$ and the maximum search depth $d=3$. This approach prevents the retriever from managing an excessive number of paths while still providing sufficient information for effective retrieval.


\subsection{Dataset Statistics}\label{ap:dataset}
Comprehensive statistics for open-domain and closed-domain QA benchmarks, including dataset splits, are presented in Table \ref{tab:dataset}.

\begin{table}[!h]
\centering
\setlength\tabcolsep{3pt}
\resizebox{0.86\linewidth}{!}{
\begin{tabular}{lcccccc}
\toprule
\textbf{Dataset} & \textbf{Train} & \textbf{Validation} & \textbf{Test} & \textbf{Total} \\
\midrule
Wikitopics & 89815 & 89726 & 89749 & 269290 \\
HotpotQA & 640 & 160 & 200 & 1,000 \\
MuSiQue & 640 & 160 & 200 & 1,000 \\
2WikiMultiHopQA & 640 & 160 & 200 & 1,000 \\
\bottomrule
\end{tabular}
}
\caption{Statistics of QA benchmarks across domain settings}\label{tab:dataset}
\end{table}

\subsection{Github Repository}
Our anonymized code is available at \url{https://github.com/Vincent-Lien/HyperRAG.git}.

\begin{figure}[!ht]
  \centering
  \begin{subfigure}[b]{\columnwidth}
    \centering
    \includegraphics[width=\columnwidth]{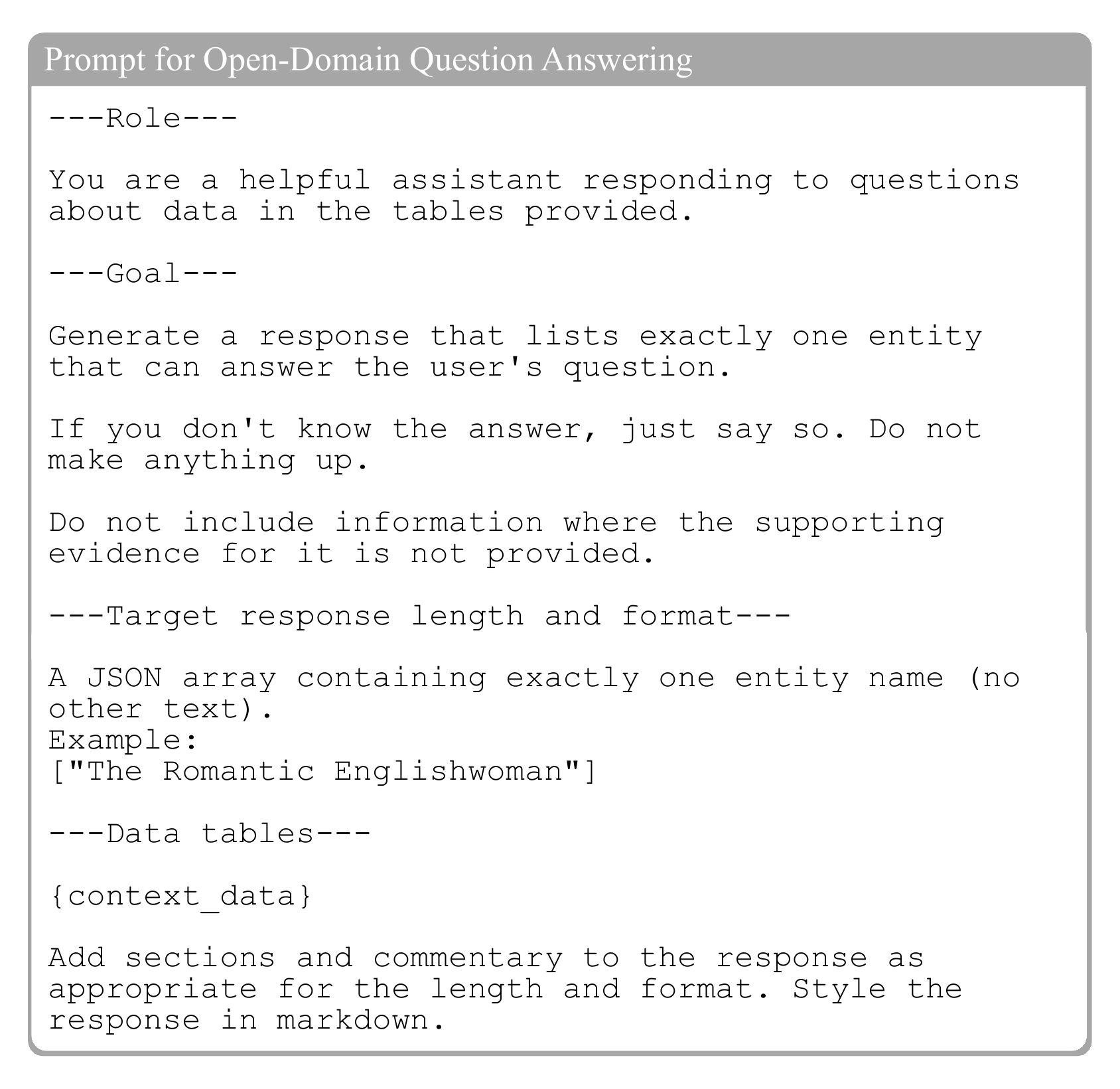}
    \caption{Open-Domain Question Answering}
    \label{fig:open-qa}
  \end{subfigure}
  
  
  \begin{subfigure}[b]{\columnwidth}
    \centering
    \includegraphics[width=\columnwidth]{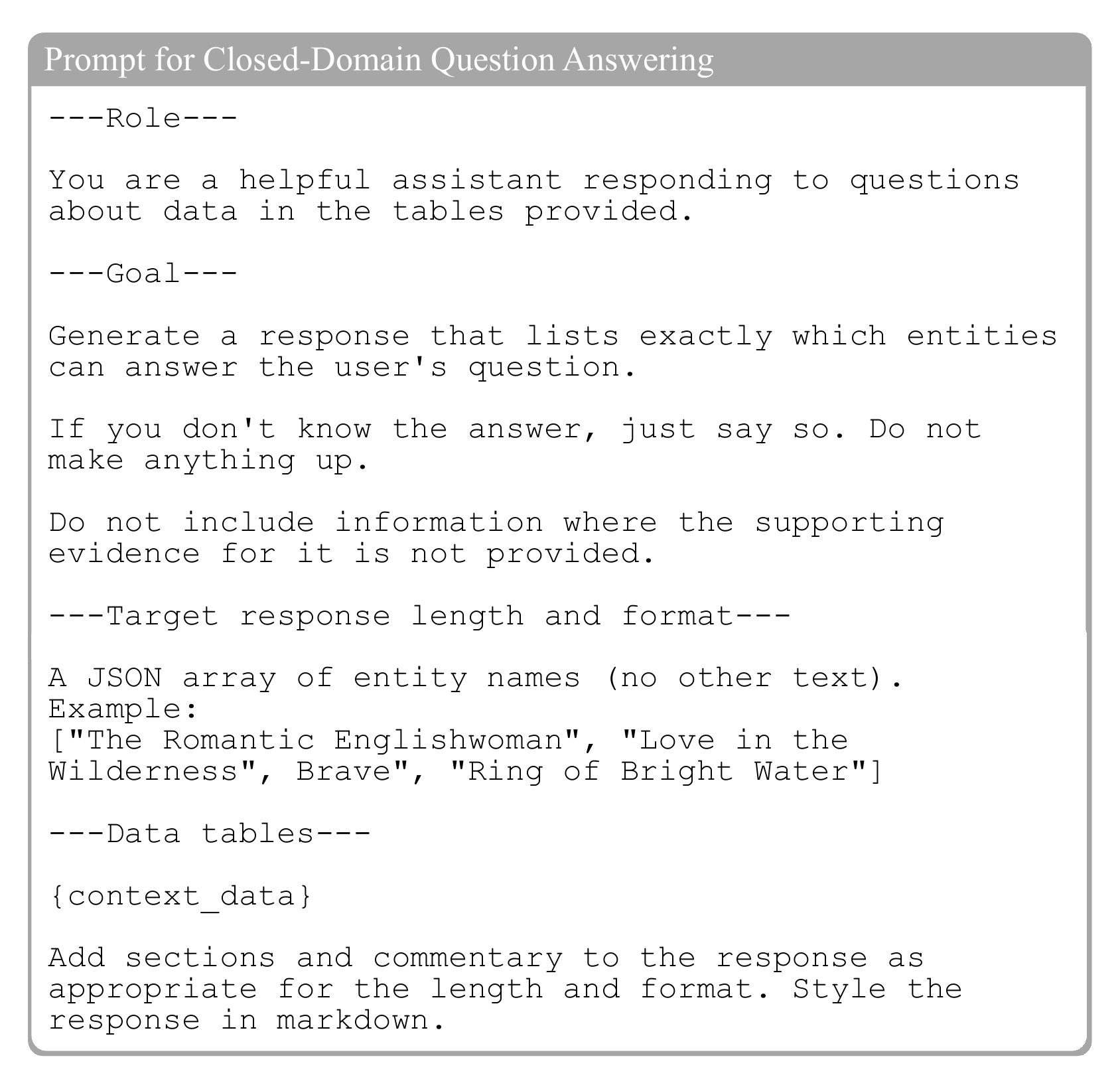}
    \caption{Closed-Domain Question Answering}
    \label{fig:close-qa}
  \end{subfigure}
  
  \caption{Prompt templates for (a) Open-Domain Question Answering, and (b) Closed-Domain Question Answering.}
  \label{fig:grid2}
\end{figure}

\balance

\section{Additional Qualitative Results}\label{ap:quality}
\begin{figure}[!b]
    \centering
    \includegraphics[width=1.0\linewidth]{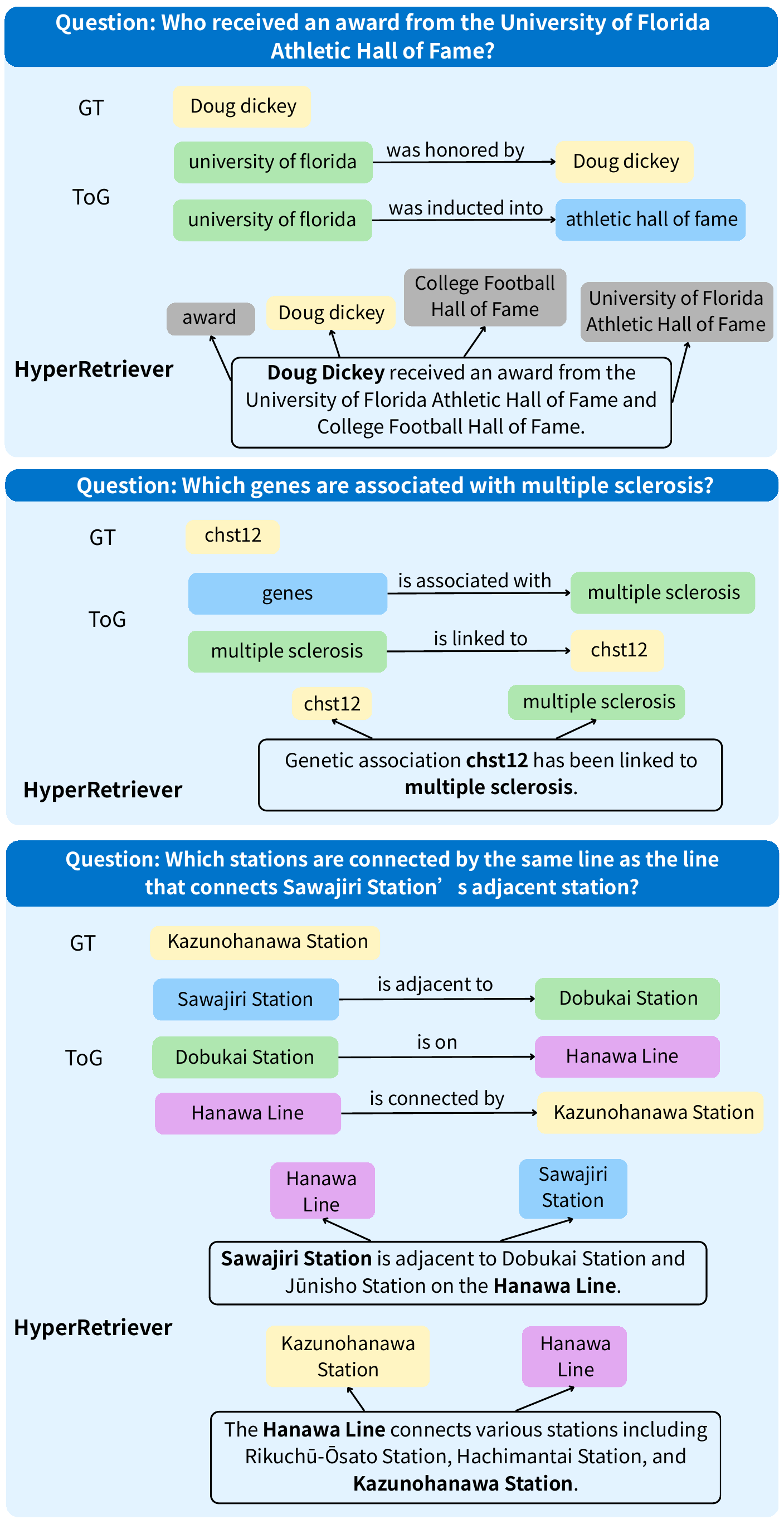}
    \caption{Comparison of evidential $n$-ary relational chains. We contrast Ground-Truth (GT) answers with reasoning paths derived by ToG and HyperRetriever. While ToG operates on standard knowledge graphs restricted to binary relations, HyperRetriever leverages hypergraphs to preserve the semantic integrity of dense $n$-ary facts.}
    \label{fig:case}
\end{figure}

Figure~\ref{fig:case} provides a qualitative comparison of evidential $n$-ary relational chains extracted by the strong baseline, ToG, versus our proposed HyperRetriever, alongside the Ground Truth (GT). The analysis reveals that HyperRetriever exploits hypergraph topology to preserve the semantic integrity of dense $n$-ary facts, resulting in structurally concise reasoning paths. Conversely, ToG is constrained by binary graph decomposition, necessitating longer, more fragmented traversal paths to capture equivalent dependencies.

\begin{figure*}[t]
  \centering
  \begin{subfigure}[t]{0.85\textwidth}
    \centering\includegraphics[width=\linewidth]{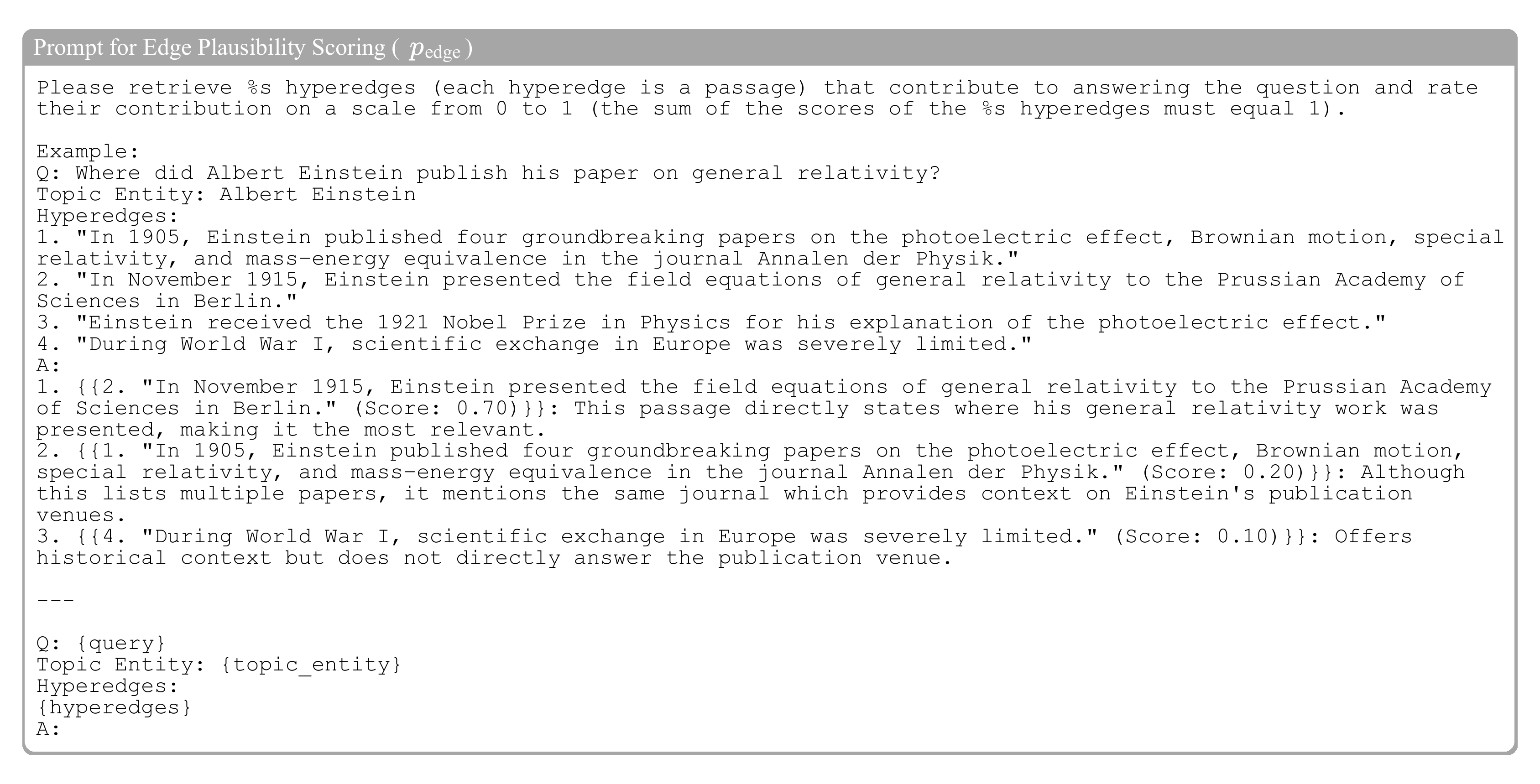}
    \caption{Edge Plausibility Scoring ($p_{\text{edge}}$)}\label{fig:plausibility}
  \end{subfigure}\hfill
  \begin{subfigure}[t]{0.85\textwidth}
    \centering\includegraphics[width=\linewidth]{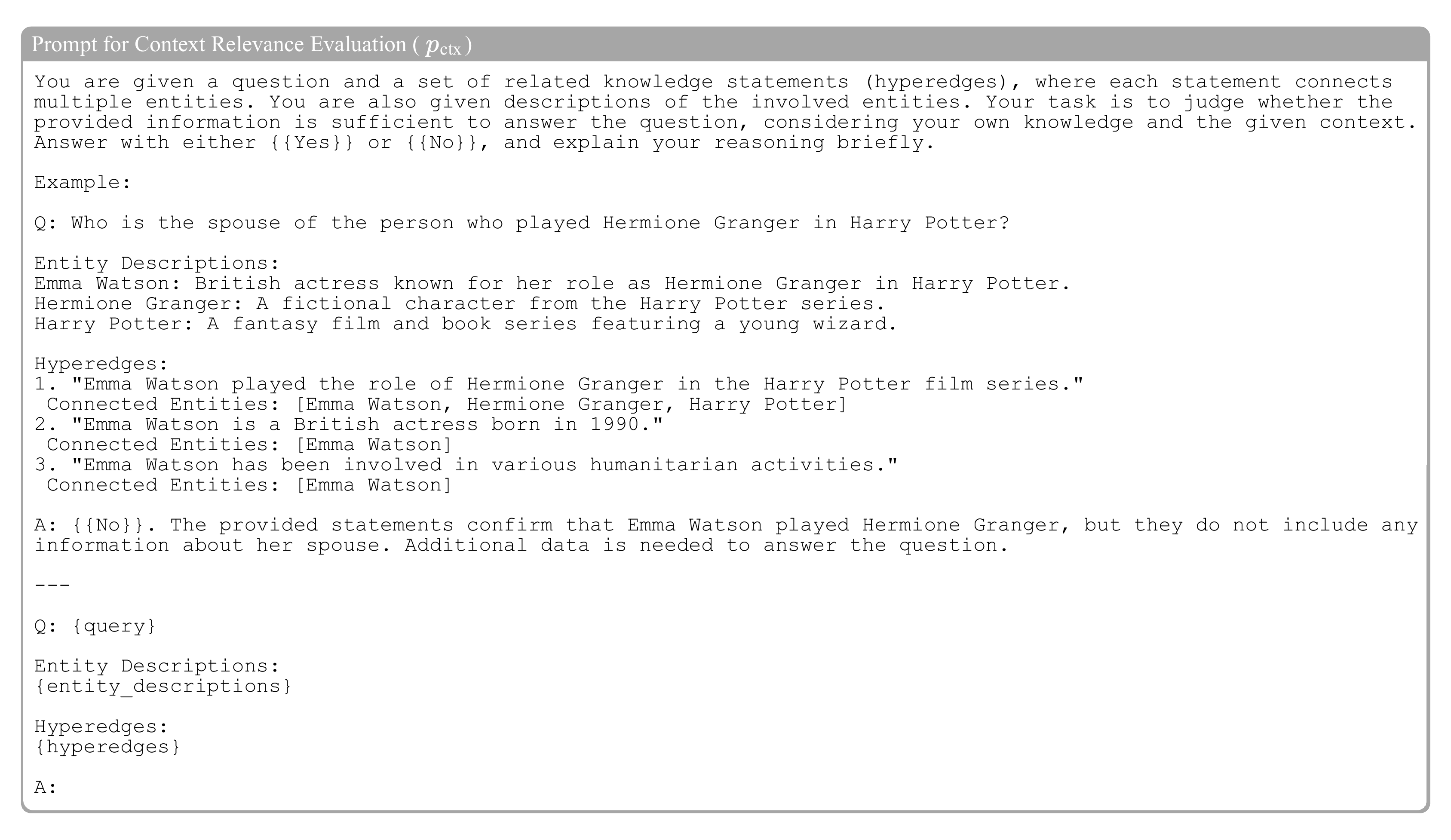}
    \caption{Context Relevance Evaluation ($p_{\text{ctx}}$)}\label{fig:context}
  \end{subfigure}
  \caption{Prompt for (a) Edge Plausibility Scoring, and (b) Context Relevance Evaluation.}
  \label{fig:grid1}
\end{figure*}




\section{Prompt Templates}\label{ap:promtp}
\noindent \textbf{Edge Plausibility Scoring (\texorpdfstring{$p_{\text{edge}}$).}{}} 
The template for $p_{\text{edge}}$ is depicted in Figure \ref{fig:plausibility}.

\noindent \textbf{Entity Salience Scoring (\texorpdfstring{$p_{\text{entity}}$}{}).}
The template for $p_{\text{edge}}$ is depicted in Figure \ref{fig:entity}.

\noindent \textbf{Context Relevance Evaluation (\texorpdfstring{$p_{\text{ctx}}$}{}).}
The template for $p_{\text{ctx}}$ is depicted in Figure \ref{fig:context}.

\noindent \textbf{Question Answering.}
We generate the final answers for both HyperRetriever and HyperMemory using the same prompt and dataset. For open-domain QA benchmarks such as HotpotQA, MuSiQue and 2WikiMultiHopQA, the answer is usually a single entity or sentence. Therefore, we design the prompt to guide the model toward a clear, single factual reply.
In contrast, the closed-domain WikiTopics-CLQA dataset expects a list of multiple entities. In this case, we shape the prompt to ensure the model produces a list of all relevant entities, thus ensuring the output matches the required multi-item format.

\par\noindent \textbf{Prompt for Open-Domain Question Answering: }
The template for open-domain question answering is illustrated in Figure \ref{fig:open-qa}.

\par\noindent \textbf{Prompt for Closed-Domain Question Answering: }
The template for closed-domain question answering is given in Figure \ref{fig:close-qa}.






%

\end{document}